\documentclass{article}
\usepackage{spconf, amsmath, amssymb, amsfonts}
\usepackage{array, multirow, color, booktabs, textcomp}
\usepackage{enumitem, amsfonts, amssymb, float}
\usepackage{xcolor, bm, comment}
\usepackage{mathtools, balance}
\usepackage{algorithmic, algorithm, caption, preview}
\usepackage{cite, hyperref, colortbl, makecell}
\usepackage{threeparttable,booktabs}
\usepackage[normalem]{ulem}
\usepackage{svg}
\usepackage{pifont, import, booktabs}
\usepackage{CJKutf8}
\usepackage{siunitx}
\usepackage{arydshln}
\usepackage[justification=centering,font=scriptsize,skip=0pt]{subcaption}
\usepackage{float}
\usepackage{placeins}
\usepackage{etoolbox}
\usepackage{orcidlink}
\apptocmd{\thebibliography}{\fontsize{9}{10}\selectfont}{}{}

\title{Memory, Know Your PLACE: Governing Lifecycle State \\ and Query-Time Eligibility in Persistent-Memory Agents}
\name{Chen-Fang Hu$^1$\orcidlink{0009-0001-0761-4676}, 
Hong-Han Shuai$^{2}$\orcidlink{0000-0003-2216-077X},
Woan-Shiuan Chien$^{2}$\orcidlink{0000-0003-2235-4080}}

\address{
    $^1$Arête Honors Program, National Yang Ming Chiao Tung University, Taiwan \\
    $^2$Institute of Electrical and Computer Engineering, National Yang Ming Chiao Tung University, Taiwan
}
\begin{document}

\maketitle
\begin{abstract}
Persistent memory turns prior interactions into durable agent state, but existing systems provide limited control over whether that state remains valid and whether it should influence a later query. We argue that reliable memory requires governance over both state formation and query-time influence. We introduce \textbf{PLACE}, a factorized governance layer that adds write-side state certification and read-side authorization to a memory backend. A factorial evaluation on Mem0 with Qwen2.5-3B and Qwen2.5-7B isolates the contribution of each mechanism. Results show that write-side governance substantially improves lifecycle state correctness, while read-side governance eliminates measured causal leakage from ineligible memory in the state-correct setting. Their combination achieves the strongest joint performance while preserving query-relevant support. These findings show that reliable persistent memory requires explicit control over both maintained state and query-time information influence.
\end{abstract}

\vspace{-1mm}
%

%%%%%%%%%%%%%%%%%%%%%%%%%%%%%%%%%%%%%%%%%%%%%%%%%%%%%%%%%%%%%%%%%%%%%%%%%%%%%%%%
% Introduction 1 page 
%%%%%%%%%%%%%%%%%%%%%%%%%%%%%%%%%%%%%%%%%%%%%%%%%%%%%%%%%%%%%%%%%%%%%%%%%%%%%%%%
\vspace{-1mm}
\section{Introduction}
\label{sec:intro}
\vspace{-1mm}
Long-running agents increasingly rely on persistent memory to maintain continuity beyond a single interaction \cite{mem0, packer2023memgpt}. Persistent memory allows agents to preserve evolving user and task states across sessions, so later responses can reflect information established in earlier interactions \cite{memorybank}. Recent memory-augmented systems have therefore moved beyond static-context formulations toward editable memory substrates that can be constructed, updated, retrieved, and reused over time \cite{mem0,Amem, rasmussen2025zep}. Evaluation has likewise moved beyond isolated recall toward long-horizon settings involving temporal reasoning, memory updates, forgetting, and state evolution \cite{longmemeval, memoryagentbench, uddin2026recall}. As persistent memory becomes more central to agent behavior, however, errors in what the agent stores or uses can also persist and affect later behavior \cite{xiong2026memory}.

Greater operational flexibility does not by itself make persistent memory governable. Prior systems improve how agents construct, update, and retrieve useful memory from earlier interactions \cite{secom, yu2026agentic}. Recent benchmarks instead examine whether memory remains useful as information evolves across long interactions \cite{amemgym, uddin2026recall}. Yet these advances do not establish whether the persistent backend itself represents the correct current semantic state. An agent may answer correctly from nearby context even when obsolete information remains active in persistent memory \cite{hao2026memops}. Conversely, it may maintain the correct current state while later exposing stale or query-ineligible memory to the answer model \cite{zhang2026beyond, cimemories}. Final-answer behavior alone therefore cannot reveal whether an error originates in memory state or in how that state is allowed to influence a query, making storage quality or answer accuracy alone insufficient to characterize memory correctness.

We argue that reliable persistent memory requires governance over two distinct but complementary problems. \textit{State correctness} requires memory to maintain the currently valid semantic state as information evolves across an interaction history. \textit{Influence correctness}, by contrast, requires ensuring that only memory eligible for the current query can affect the response, while preserving access to relevant and authorized memory. This distinction leads to two research questions:
\begin{itemize}[leftmargin=*,topsep=0pt,itemsep=0pt,parsep=0pt,partopsep=0pt]
\item RQ1: Can write-side governance improve lifecycle state consistency during memory updates?
\item RQ2: Can read-side governance prevent query-ineligible memory from influencing responses while preserving query-relevant memory utility?
\end{itemize}

To address these questions, we propose \textbf{PLACE}, a factorized governance layer for persistent-memory agents. PLACE adds explicit control over how memory is formed and how represented memory is allowed to influence downstream behavior. Specifically, its write-side mechanism maintains and verifies the semantic state physically realized in the backend, while its read-side mechanism exposes only support authorized for the current query. We evaluate PLACE on Mem0 \cite{mem0} with Qwen2.5-3B and Qwen2.5-7B \cite{qwen25} in a factorial design that isolates write-side and read-side effects. 
%Across both model scales, write-side governance keeps the realized memory state correct throughout the evaluation, while read-side governance eliminates observed causal leakage from ineligible memory when the stored state is correct. 
%Combining the two yields the strongest overall performance, showing that reliable agent memory requires controlling both what persistent memory becomes and what it is allowed to influence. 
Across both model scales, write-side governance maintains a correct memory state, while read-side governance eliminates observed causal leakage when the state is correct. Combining both yields the strongest overall performance, showing that reliable agent memory requires controlling both memory state and influence. Project materials are available online\footnotemark.

\begin{figure*}[!t]
    \centering
    \includegraphics[width=\textwidth]{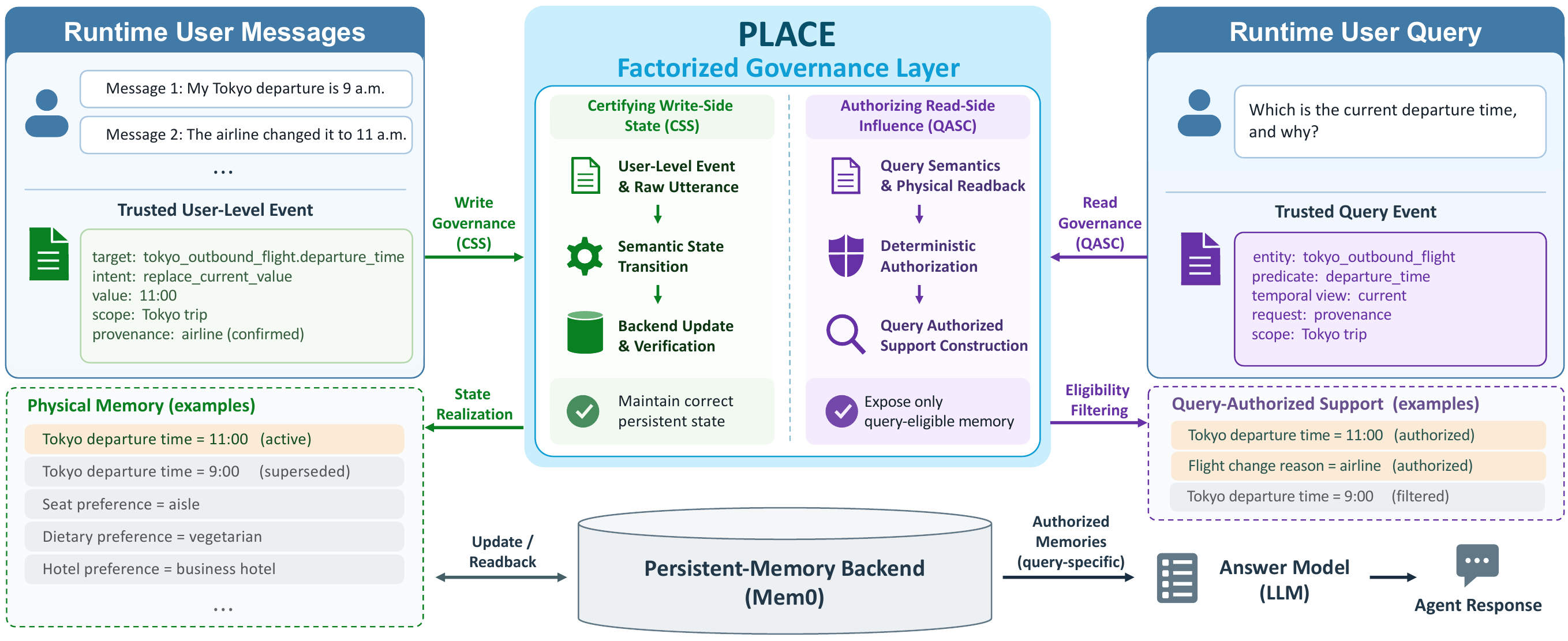}
%    \caption{PLACE separates memory governance into write-side state correctness and read-side influence correctness. CSS governs and certifies persistent-state formation, whereas QASC exposes only query-eligible support already represented in physical memory. The flight example illustrates the distinction: CSS replaces the current departure value, while QASC prevents the superseded value from influencing a current-time query.}
\caption{Overview of the proposed PLACE framework, which separates persistent-memory governance into write-side state correctness and read-side influence correctness. CSS governs and certifies state formation in physical memory, while QASC exposes only query-eligible support to the answer model. The flight example illustrates the two controls by updating the active departure time and blocking the superseded value from influencing the current query.}

    \label{fig:framework}
\end{figure*}

\section{Methodology}
\label{ssec:method}
\vspace{-2mm}
\subsection{Problem Formulation}
\label{ssec:problem-formulation}
\vspace{-2mm}
We formulate persistent-memory governance as two distinct control problems: maintaining the correct semantic state in physical storage and controlling which part of that state may influence a query. Let $X_t$ denote the physical memory backend after interaction step $t$. Its evaluator-side semantic readback is defined as 
{\setlength\abovedisplayskip{1mm}
\setlength\belowdisplayskip{1mm}
\begin{equation}
    S_t^{\mathrm{phys}} = \beta(X_t),
    \label{eq:physical-readback}
\end{equation}}
where $\beta$ maps only physically evidenced records to semantic facts and cannot introduce unsupported values. The reference state $S_t^{\star}$ is used only for evaluation and is never exposed to the runtime system. Write-side correctness asks whether $S_t^{\mathrm{phys}}$ remains consistent with $S_t^{\star}$ as memory evolves. Missing replacements, stale values, and contradictions therefore remain visible as state errors even when the final response is correct.

Read-side correctness asks whether represented memory is eligible to influence the current response. Let $R_t$ denote the physical memory view available to the read-side policy, and let $\tau_t$ denote the trusted interpretation of query $q_t$. We define the authorized support as
{\setlength\abovedisplayskip{1mm}
\setlength\belowdisplayskip{1mm}
\begin{equation}
    Q_t =
    \left\{
    m \in R_t
    \mid
    E(m,q_t,\tau_t)=1
    \right\},
    \label{eq:qasc-authorize}
\end{equation}}
where $E$ is the eligibility rule and $Q_t$ is the only memory support exposed to the answer model. Because $Q_t \subseteq R_t$, read-side governance can filter represented information but cannot reconstruct facts absent from physical memory. 

\footnotetext{\scriptsize\url{https://hcf-tracy.github.io/memory-know-your-place/}}

% ============================================================
% V32 / v3.2-strict-general evaluator
% ============================================================

\begin{table*}[t]
\centering
%\footnotesize

\normalsize

%\caption{Primary factorial results and prespecified governance effects. Panel A reports state correctness, causal leakage, answer correctness, and joint success. Panel B summarizes the corresponding write-side, read-side, joint, and interaction effects.}
\caption{Summary of the primary factorial results and prespecified governance effects. Panel A reports the factorial outcomes, while Panel B summarizes the corresponding governance effects.}
\vspace{-2mm}
\label{tab:main-results}

\renewcommand{\arraystretch}{1.04}

% ============================================================
% PANELS A + B: KEEP ORIGINAL SIDE-BY-SIDE FORMAT
% ============================================================
\setlength{\tabcolsep}{2.2pt}

\begin{tabular*}{\textwidth}{
@{\extracolsep{\fill}}
l@{\hspace{1pt}}lcccc
@{\hspace{3pt}}
l@{\hspace{1pt}}ll
@{}
}
\toprule

\multicolumn{6}{l}{\textbf{A. Primary factorial}}
&
\multicolumn{3}{l}{\textbf{B. Prespecified effects}}
\\

\cmidrule(lr){1-6}
\cmidrule(lr){7-9}

Model
& Cond.
& \shortstack{State\\Correctness$\uparrow$}
& \shortstack{Leakage\\Rate$\downarrow$ $(n)$}
& \shortstack{Answer\\Correctness$\uparrow$}
& \shortstack{Joint\\Success$\uparrow$}
& Model
& Effect
& Estimate [interval]
\\
\midrule

7B
& Native
& 0.488
& 0.071 (28)
& 0.682
& 0.000
& 7B
& StateGain
& 0.512 [0.447, 0.599]
\\

& CSS
& \textbf{1.000}
& 0.125 (48)
& \textbf{0.793}
& 0.417
& 
& LeakageDrop
& 0.125 [0.042, 0.229]
\\

& QASC
& 0.488
& 0.000 (28)
& 0.653
& 0.000
& 
& JointGain
& 0.667 [0.610, 0.964]
\\

& \textbf{PLACE}
& \textbf{1.000}
& \textbf{0.000 (48)}
& 0.713
& \textbf{0.667}
& 
& InteractionGain
& 0.250 [0.060, 0.377]
\\

\addlinespace[1pt]
\cmidrule(lr){1-6}
\cmidrule(lr){7-9}

3B
& Native
& 0.414
& 0.071 (14)
& 0.451
& 0.000
& 3B
& StateGain
& 0.586 [0.539, 0.660]
\\

& CSS
& \textbf{1.000}
& 0.083 (48)
& 0.618
& 0.354
& 
& LeakageDrop
& 0.083 [0.021, 0.167]
\\

& QASC
& 0.414
& 0.000 (14)
& 0.410
& 0.000
& 
& JointGain
& 0.625 [0.592, 0.917]
\\

& \textbf{PLACE}
& \textbf{1.000}
& \textbf{0.000 (48)}
& \textbf{0.654}
& \textbf{0.625}
& 
& InteractionGain
& 0.271 [0.069, 0.377]
\\

\bottomrule
\end{tabular*}

\end{table*}

\subsection{Overall Framework} 
\label{ssec:overall-framework} 
\vspace{-1mm}
\textbf{PLACE} (Persistent-memory Lifecycle Alignment and Contextual Eligibility) adds governance at the write and read boundaries of an existing persistent-memory system, as shown in Fig.~\ref{fig:framework}, while leaving the memory backend and answer model unchanged. CSS governs write-side state formation by applying trusted user-level events, realizing the resulting state in memory, and verifying it through physical readback. QASC governs read-side influence by exposing only query-eligible support to the answer model. Together, both mechanisms address complementary failures. CSS prevents stale or conflicting states from persisting, while QASC prevents ineligible memory from influencing the response.

\vspace{-2mm}
\subsubsection{CSS: Certifying Write-Side State}
\label{sssec:module-1}
\vspace{-2mm}
Persistent update errors documented in prior work motivate CSS to explicitly govern state transitions \cite{yang2026trustmem, cui2026memtxn}. Given the previous governed state $G_{t-1}$ and a trusted user-level event $e_t$, it computes
{\setlength\abovedisplayskip{0mm}
\setlength\belowdisplayskip{0mm}
\begin{equation}
    G_t = F_{\mathrm{CSS}}(G_{t-1},e_t).
    \label{eq:css-transition}
\end{equation}}
The trusted event encodes user-level update intent, such as the target fact, its value, and scope, but not the resulting lifecycle state or internal CSS transition. CSS then derives the required state change, realizes $G_t$ through backend operations, and verifies resulting physical state through Eq.~\ref{eq:physical-readback}.

\vspace{-1mm}
\subsubsection{QASC: Authorizing Read-Side Influence}
\label{sssec:module-2}
\vspace{-1mm}
Prior evidence motivates query-time eligibility beyond semantic relevance \cite{zhang2026beyond, cimemories, zhan2026memory}. QASC then enforces this criterion over physically represented memory. The eligibility rule $E$ considers the requested entity and predicate together with relevant scope, temporal, and lifecycle constraints. Only items satisfying Eq.~\ref{eq:qasc-authorize} enter the model-visible support. Moreover, QASC is read-only and grounded in physical storage. It cannot modify persistent state or recover information absent from the backend. Authorization precedes context-budget truncation, preventing an ineligible item from reaching the answer model merely because it ranks highly under retrieval.

%%%%%%%%%%%%%%%%%%%%%%%%%%%%%%%%%%%%%%%%%%%%%%%%%%%%%%%%%%%%%%%%%%%%%%%%%%%%%%%%
% Experimental Setup and Results
%%%%%%%%%%%%%%%%%%%%%%%%%%%%%%%%%%%%%%%%%%%%%%%%%%%%%%%%%%%%%%%%%%%%%%%%%%%%%%%%
\vspace{-1mm}
\section{Experimental Setup and Results}
\label{ssec:exp}
\vspace{-1mm}
\subsection{Experimental Design}
\label{ssec:experimental-design}
\vspace{-1mm}
\textbf{\textit{Implementation.}} We evaluate PLACE on Mem0 \cite{mem0} with Qwen2.5-3B and Qwen2.5-7B \cite{qwen25} through Ollama, using Qdrant storage and Nomic embeddings. The same model drives Native memory updates and answer generation, while CSS and QASC are deterministic.

\noindent\textbf{\textit{Factorial design.}} We cross Native and CSS writing with Native and QASC reading, yielding Native, CSS, QASC, and PLACE. Native and QASC share the same Native-produced state, while CSS and PLACE share the same CSS-produced state. Each producer is written once per update step to isolate read-side effects.

\noindent\textbf{\textit{Primary evaluation.}} Experiments use one prespecified 302-turn trajectory spanning four conversations, with persistent memory retained across session resets. It contains 105 state checkpoints and 48 queries across six governance families. Primary uses trusted user-level event and query semantics, without exposing the resulting lifecycle state, internal CSS transition, or authorized support. Read-side influence is further isolated with 48 causal pairs that hold the query and authorized state fixed while varying only irrelevant history.

\noindent\textbf{\textit{Diagnostics.}} Utility evaluates needed support, inappropriate exposure, and conflicting evidence over 72 probes. Closed Loop repeats the factorial setting while retaining prior assistant responses as non-authoritative dialogue context.

\subsection{Metrics and Statistical Analysis}
\label{ssec:metrics}

\noindent\textbf{\textit{State correctness.}} Write-side governance is evaluated from the physical backend state defined in Eq.~\ref{eq:physical-readback}. Across the $C=105$ state checkpoints, we compute
{\setlength\abovedisplayskip{1mm}
\setlength\belowdisplayskip{1mm}
{\small
\begin{equation}
\mathrm{StateCorrectness}
=\frac{1}{C}
\sum_{c=1}^{C}
F_1\!\left(S_c^{\mathrm{phys}},S_c^{\star}\right).
\label{eq:state-correctness}
\end{equation}}}
This metric evaluates the state actually realized in storage rather than an intended internal representation. The checkpoints form one longitudinal trajectory and are therefore not treated as independent experimental units.

\noindent\textbf{\textit{Factorization check.}} Before comparing read policies, we verify that paired conditions observe identical producer states. Native must match QASC, and CSS must match PLACE, under the same semantic readback at all 105 post-update and 48 pre-query checkpoints. Any mismatch would invalidate the corresponding read-side comparison.

\noindent\textbf{\textit{Causal leakage.}} Read-side influence is measured from paired worlds described above. For world $w\in\{a,b\}$, let $Q_{\mathrm{eval}}^{(w)}$ denote the evaluator-authorized state derived from physical memory. Leakage is evaluated only when both worlds realize the same reference-correct authorized state $Q^{\star}$,
{\setlength\abovedisplayskip{1mm}
\setlength\belowdisplayskip{1mm}
{\small
\begin{equation}
\mathrm{LeakageRate}
=
\Pr\!\left(
\Omega^{(a)}\neq\Omega^{(b)}
\mid
Q_{\mathrm{eval}}^{(a)}
=
Q_{\mathrm{eval}}^{(b)}
=
Q^{\star}
\right)
\label{eq:leakage-rate}
\end{equation}}}
where $\Omega^{(w)}$ denotes the semantic answer outcome. Pairs with different authorized states are classified as state-mediated and excluded from the leakage analysis, while equal but reference-incorrect pairs are excluded from the denominator.

\noindent\textbf{\textit{Answer and joint correctness.}} AnswerCorrectness is deterministic semantic F1 over generated answers. JointSuccess is stricter and requires the persistent state and authorized support to be correct, required support to remain available, no governance violation to occur, and the final answer to be correct. The interaction between CSS and QASC is summarized by a factorial difference-in-differences on JointSuccess.

\noindent\textbf{\textit{Utility diagnostics.}} NeededSupport measures whether query-required memory remains visible to the answer model, while ExposureRate measures whether ineligible memory is exposed. ConflictHarm records whether injected conflicting evidence contributes to an incorrect answer. Closed Loop reports the same state, exposure, and joint endpoints when prior assistant responses remain in dialogue context.

\noindent\textbf{\textit{Uncertainty.}} Because the primary benchmark contains one prespecified four-conversation trajectory, longitudinal results are reported descriptively with conversation-block sensitivity intervals rather than population-level inference. Uncertainty for the 48 independent causal pairs is estimated with 4,000 pair-level bootstrap resamples. Results from Qwen2.5-3B and Qwen2.5-7B are treated as model-scale replications.

% ============================================================
% SINGLE-COLUMN TARGETED DIAGNOSTICS
% ============================================================

\begin{table}[t!]
\centering
%\scriptsize
\normalsize

\caption{Summary of targeted diagnostics for Utility and Closed Loop, covering support preservation, exposure control, and joint governance performance.}
\vspace{-2mm}
\label{tab:targeted-diagnostics}

\setlength{\tabcolsep}{1.8pt}
\renewcommand{\arraystretch}{0.96}

\begin{tabular*}{\columnwidth}{
@{}
p{0.17\columnwidth}
l
l
@{\extracolsep{\fill}}
c
c
c
@{}
}
\toprule

&
Model
& Cond.
& \shortstack{Needed\\Support$\uparrow$}
& \shortstack{Exposure\\Rate$\downarrow$}
& \shortstack{Conflict\\Harm$\downarrow$}
\\
\midrule

\multirow{8}{0.17\columnwidth}{\centering \textbf{A. Read-side Utility}}
& 7B
& Native
& 0.200
& 0.292
& 0.125
\\

& 
& CSS
& 0.667
& 0.417
& \textbf{0.000}
\\

& 
& QASC
& 0.200
& 0.000
& 0.042
\\

& 
& \textbf{PLACE}
& \textbf{1.000}
& \textbf{0.000}
& 0.042
\\

\cmidrule(lr){2-6}

& 3B
& Native
& 0.133
& 0.250
& 0.042
\\

& 
& CSS
& 0.667
& 0.417
& \textbf{0.000}
\\

& 
& QASC
& 0.133
& 0.000
& \textbf{0.000}
\\

& 
& \textbf{PLACE}
& \textbf{1.000}
& \textbf{0.000}
& \textbf{0.000}
\\

% ------------------------------------------------------------
% Closed Loop
% ------------------------------------------------------------

\midrule

&
Model
& Cond.
& \shortstack{State\\Correctness$\uparrow$}
& \shortstack{Exposure\\Rate$\downarrow$}
& \shortstack{Joint\\Success$\uparrow$}
\\
\midrule

\multirow{8}{0.17\columnwidth}{\centering \textbf{B. Closed Loop}}
& 7B
& Native
& 0.488
& 0.292
& 0.000
\\

& 
& CSS
& \textbf{1.000}
& 0.354
& 0.417
\\

& 
& QASC
& 0.488
& 0.000
& 0.000
\\

& 
& \textbf{PLACE}
& \textbf{1.000}
& \textbf{0.000}
& \textbf{0.667}
\\

\cmidrule(lr){2-6}

& 3B
& Native
& 0.418
& 0.229
& 0.000
\\

& 
& CSS
& \textbf{1.000}
& 0.375
& 0.354
\\

& 
& QASC
& 0.418
& 0.000
& 0.000
\\

& 
& \textbf{PLACE}
& \textbf{1.000}
& \textbf{0.000}
& \textbf{0.458}
\\

\bottomrule
\end{tabular*}
\end{table}

\vspace{-2mm}
\subsection{Results and Discussion}
\label{ssec:results}
\vspace{-1mm}
\subsubsection{Write-Side Governance Improves State Correctness}
\vspace{-1mm}
Table~\ref{tab:main-results} shows a clear write-side effect. StateCorrectness increases from 0.488 to 1.000 at 7B and from 0.414 to 1.000 at 3B, corresponding to StateGain values of 0.512 and 0.586, respectively. QASC does not change these values because it shares the Native state producer and does not modify persistent memory. The factorization check also holds across all 153 semantic checkpoints, with Native matching QASC and CSS matching PLACE at both model scales. These results indicate that the write-side gain reflects improved state formation rather than differences introduced at retrieval time.

\subsubsection{Read-Side Governance Reduces Ineligible Influence}
%\vspace{-1mm}
CSS and PLACE provide the cleanest read-side comparison because both realize the reference-correct state for all 48 causal pairs and differ only in their read policy. LeakageRate decreases from 0.125 to 0.000 at 7B and from 0.083 to 0.000 at 3B, with 95\% pair-bootstrap intervals of $[0.042,0.229]$ and $[0.021,0.167]$ for LeakageDrop. The smaller eligible sets for Native and QASC arise from write-side state errors and are therefore not used as the primary read-side contrast.

As shown in Table~\ref{tab:targeted-diagnostics}, Utility diagnostic shows that the reduction in leakage is not achieved by indiscriminate suppression. PLACE preserves 1.000 NeededSupport at both model scales while reducing ExposureRate to 0.000. Read-side authorization therefore removes measured ineligible influence while preserving all measured query-required support.
%Read-side authorization therefore removes measured ineligible influence while retaining most memory required by the query.

\subsubsection{Combining Both Controls Improves Joint Performance}

As shown in Table~\ref{tab:main-results}, CSS and QASC are most effective when combined. PLACE reaches a JointSuccess of 0.667 at 7B and 0.625 at 3B, compared with 0.417 and 0.354 for CSS, respectively, while QASC alone cannot recover failures caused by an incorrect Native state. InteractionGain is 0.250 at 7B and 0.271 at 3B, with sensitivity intervals of $[0.060,0.377]$ and $[0.069,0.377]$. We therefore interpret the interaction as evidence consistent with complementarity rather than as population-level statistical synergy.

Table~\ref{tab:targeted-diagnostics} further shows that this pattern persists in the Closed-Loop setting. PLACE maintains StateCorrectness of 1.000 and ExposureRate of 0.000 at both model scales, with JointSuccess of 0.667 at 7B and 0.458 at 3B.  These results indicate that the combined governance remains effective even when prior assistant responses are retained as non-authoritative dialogue context.

%CSS and QASC mechanisms are most effective when combined. PLACE reaches a JointSuccess of 0.542 at both model scales, compared with 0.375 and 0.333 for CSS at 7B and 3B, respectively, while QASC alone cannot recover failures caused by an incorrect Native state. This pattern supports the intended separation between the two governance planes. Read-side authorization can control which represented memories are used, but it cannot repair information that was incorrectly formed or is absent from storage.

%Table~\ref{tab:targeted-diagnostics} also shows that the same pattern persists in the Closed-Loop setting. PLACE maintains StateCorrectness of 1.000 and ExposureRate of 0.000 at both model scales, with JointSuccess of 0.479 at 7B and 0.438 at 3B. InteractionGain is positive at both scales, although its sensitivity interval crosses zero at 7B. We therefore interpret the factorial interaction as evidence of complementary mechanisms rather than a scale-uniform statistical synergy.

%%%%%%%%%%%%%%%%%%%%%%%%%%%%%%%%%%%%%%%%%%%%%%%%%%%%%%%%%%%%%%%%%%%%%%%%%%%%%%%%
% Conclusion
%%%%%%%%%%%%%%%%%%%%%%%%%%%%%%%%%%%%%%%%%%%%%%%%%%%%%%%%%%%%%%%%%%%%%%%%%%%%%%%%
%\vspace{-1mm}
\section{Conclusion and Future Work}
\label{sec:conclusion}
In this work, we introduce PLACE, a factorized governance layer that separates persistent-state formation from query-time memory influence. Across Qwen2.5-3B and Qwen2.5-7B, write-side governance achieves perfect StateCorrectness, while read-side authorization eliminates measured causal leakage in the state-correct setting. Their combination yields the strongest joint performance while preserving query-needed support. These results highlight two core contributions: 1) state correctness and influence correctness are distinct but complementary requirements for reliable persistent memory; and 2) improving retrieval alone is insufficient when the underlying persistent state is incorrect. Our current evaluation isolates governance using trusted user-level semantics rather than end-to-end automatic interpretation. Future work will therefore extend PLACE with learned semantic interpretation and evaluate it across independently replicated long-horizon trajectories with human validation.

\clearpage
\section{Acknowledgments}
\vspace{-2mm}
Generative AI tools were used for language editing, stylistic refinement, and manuscript organization only. All technical content and conclusions were developed and verified by the authors. The authors take full responsibility for the paper.
\vspace{-2mm}

%\renewcommand{\normalsize}{\fontsize{9}{10}\selectfont}
%\normalsize
% -------------------------------------------------------------------------
\bibliographystyle{IEEEbib}
\bibliography{mybib}

\end{document}